\documentclass[12pt]{article}

\usepackage{sbc-template}
\usepackage{graphicx,url}
\usepackage[utf8]{inputenc}
\usepackage[english]{babel}
\usepackage[T1]{fontenc}  
\usepackage{amsmath}   % para ambientes matemáticos como align
\usepackage{amssymb}   % para símbolos extras, incluindo \mathbb
\usepackage{amsfonts}  % para fontes matemáticas como \mathbb
\usepackage{float}
\usepackage[nolist]{acronym}
\usepackage{tikz}
\usetikzlibrary{shapes, arrows, positioning, shadows, calc}

\usepackage{subfig} % For subfigures
\usepackage[figurename=Figure,tablename=Table]{caption}

\usepackage{tikz}
\usepackage{booktabs}
\usepackage{fontawesome}

\usepackage{algorithm} 
\usepackage{algorithmic}

\title{PRISM: Perinuclear Ring-based Image Segmentation Method for Acute Lymphoblastic Leukemia Classification}

\author{Larissa Ferreira {Rodrigues Moreira}\inst{1}, Leonardo Gabriel Ferreira Rodrigues\inst{2}, \\Rodrigo Moreira\inst{1}, André Ricardo Backes\inst{3}}

\address{Institute of Exact and Technological Sciences -- Federal University of Viçosa
  (UFV)\\
  Rio Paranaíba -- MG -- Brazil
\nextinstitute
  School of Computer Science -- Federal University of Uberlândia (UFU)\\
  Uberlândia -- MG -- Brazil
  \nextinstitute
  Departament of Computing -- Federal University of São Carlos (UFSCar)\\
  São Carlos -- SP -- Brazil
  \email{\{larissa.f.rodrigues, rodrigo\}@ufv.br} \email{leonardo.g.rodrigues@ufu.br, andrebackes@ufscar.br}  
}

\begin{document} 
%-----A-----
\acrodef{AI}{Artificial Intelligence}
\acrodef{ALL}{Acute Lymphoblastic Leukemia}
%-----B-----
%-----C-----
\acrodef{CAD}{Computer-Aided Diagnosis}
\acrodef{CNN}{Convolutional Neural Network}
%-----D-----
\acrodef{DL}{Deep Learning}
%-----E-----
\acrodef{ET}{Extra Trees}
%-----F-----
\acrodef{FL}{Federated Learning}
%-----G-----
\acrodef{GLCM}{Gray Level Co-occurrence Matrix}
%-----H-----
%-----I-----
%-----J-----
%-----K-----
\acrodef{KNN}{K-Nearest Neighbors}
%-----L-----
\acrodef{LR}{Learning Rate}
\acrodef{LBP}{Local Binary Pattern}
\acrodef{LogReg}{Logistic Regression}
%-----M-----
\acrodef{MAE}{Mean Absolute Error}
\acrodef{ML}{Machine Learning}
\acrodef{MCC}{Matthews Correlation Coefficient}
%-----N-----
%-----O-----
\acrodef{OOF}{Out-Of-Fold}
%-----P-----
\acrodef{PRISM}{\textbf{P}erinuclear \textbf{R}ing-based \textbf{I}mage \textbf{S}egmentation \textbf{M}ethod}
\acrodef{PBS}{peripheral blood smears}
%-----Q-----
%-----R-----
\acrodef{RF}{Random Forest}
\acrodef{RGB}{Red, Blue, Green}
\acrodef{RBF}{Radial Basis Function}
%-----S-----
\acrodef{SVM}{Support Vector Machine}
%-----T-----
%-----U-----

%-----V-----
\acrodef{ViT}{Vision Transformer}

%-----W-----
%-----X-----
%-----Y-----
%-----Z-----

\maketitle

\begin{abstract} 
Automated analysis of peripheral blood smears for Acute Lymphoblastic Leukemia (ALL) is hindered by low contrast and substantial variability in cytoplasmic appearance, which complicate conventional membrane-based segmentation. We found that many recent approaches rely on heavy neural architectures and extensive training, but still struggle to generalize across staining and acquisition variability. To address these limitations, we propose the \textbf{P}erinuclear \textbf{R}ing-based \textbf{I}mage \textbf{S}egmentation \textbf{Me}thod (PRISM), which replaces explicit cytoplasmic delineation with adaptive concentric zones constructed around the nucleus. These perinuclear regions enable the extraction of robust cytoplasmic descriptors by integrating color information with texture statistics derived from grey-level co-occurrence patterns, without requiring accurate cell-boundary detection. A calibrated stacking ensemble of traditional classifiers leverages these descriptors to achieve a high performance, with an accuracy of 98.46\% and a precision-recall AUC of 0.9937.
\end{abstract}

\section{Introduction}\label{sec:introduction}

\ac{ALL} is an aggressive hematological malignancy marked by the rapid proliferation of immature lymphoblasts, requiring timely and accurate diagnosis to guide effective treatment. Although morphological assessment of peripheral blood smears under light microscopy remains the clinical standard, this manual procedure is labor-intensive, subjective, and constrained by the availability of trained hematologists. \ac{CAD} systems have been proposed to mitigate these limitations, yet extracting key morphological biomarkers such as nuclear enlargement and cytoplasmic reduction remains computationally challenging. Conventional \ac{CAD} pipelines rely heavily on precise cell membrane segmentation, a step that frequently fails in routine laboratory conditions because of overlapping erythrocytes, low nucleocytoplasmic contrast, and staining inconsistencies \cite{Zolfaghari2022, Kizi2025}.

Recent advances in end-to-end \ac{DL}, particularly in the field \acp{CNN} and \acp{ViT}, have shifted the focus toward segmentation-free classification \cite{Kizi2025}. Despite their strong performance on curated datasets, their clinical adoption is limited by several factors. These models lack transparent decision-making processes, hindering their acceptance in medical environments where interpretability is essential \cite{Aby2024}. Furthermore, \ac{DL} systems trained on small medical datasets are prone to overfitting and may inadvertently learn spurious correlations, such as background staining artifacts, rather than true cytological features. Their computational requirements for training and inference also restrict their deployment in resource-limited or decentralized healthcare settings. These challenges highlight the need for diagnostic frameworks that balance accuracy, interpretability, and computational efficiency \cite{RodriguesMoreira2025}.

To bridge this gap, this paper proposes the \ac{PRISM}, a method for hematological feature extraction. Instead of attempting the highly error-prone task of fully segmenting the cell membrane, \ac{PRISM} leverages the biologically stable basophilic nucleus as a geometric anchor to dynamically generate concentric, adaptive expansion rings. This approach extracts localized morphological, chromatic, and textural gradients from the immediate perinuclear interface and distal cytoplasm, transforming complex cellular pathology into a lightweight, highly discriminative tabular format that systematically rejects the background noise. By coupling this targeted feature extraction with a calibrated ensemble stacking architecture, the proposed approach achieves state-of-the-art diagnostic performance using a fraction of the computational resources. The main contributions of this study are as follows:

\begin{itemize}
    \item We propose the \ac{PRISM}, an adaptive mathematical methodology that circumvents the classic bottleneck of full-cell segmentation by analyzing isolated, multiscale perinuclear rings.
    \item Unlike opaque \ac{DL} models, \ac{PRISM} grounds its predictions in explicit, clinically verifiable morphological and chromatic gradients at the nucleocytoplasmic interface.
    \item Experimental validation demonstrating that the \ac{PRISM}-driven ensemble successfully outperforms state-of-the-art end-to-end \ac{DL} architectures on the ALL-IDB2 dataset.
    \item The translation of a high-dimensional computer vision task into a lightweight tabular machine learning problem, ensuring low-latency inference suitable for CPU-only, resource-constrained clinical environments.
\end{itemize}

The remainder of this paper is organized as follows. Section \ref{sec:related_work} provides a review of related studies on automated hematological screening and highlights current methodological gaps. Section \ref{sec:method} details the materials and proposed \ac{PRISM} method. Section \ref{sec:results_and_discussion} presents and discusses the results. Finally, Section \ref{sec:conclusion} concludes the paper and outlines the future research directions.

\section{Related Work}\label{sec:related_work}

The automated classification of \ac{ALL} has been dominated by two paradigms: handcrafted feature engineering and end-to-end \ac{DL} architectures. Early \ac{CAD} systems prioritized geometric and color properties, such as the nucleocytoplasmic ratio \cite{Scotti2005}. Subsequent studies explored keypoint descriptors \cite{Faria2018}, chromatic analysis in the HSV space \cite{Rodrigues2016}, and texture-based descriptors combined with XGBoost \cite{Dias2021}. While interpretable, these classic methods often struggle with the morphological variability of leukemic cells \cite{Zolfaghari2022}.

Conversely, \ac{DL} approaches have achieved high accuracy by leveraging complex architectures \cite{Aby2024, Kizi2025}. Recent works include hybrid ensembles of MobileNetV2, ResNet, and ShuffleNet \cite{Das2021, Das2025}, as well as Bayesian \acp{CNN} for uncertainty estimation \cite{Hita2026}. To optimize these models, meta-heuristic strategies such as Elephant Herd \cite{Sahlol2017} and Salp Swarm \cite{Sahlol2020} algorithms have been employed. \cite{Rodrigues2022} attempted to refine this process using a Genetic Algorithm (GA) to optimize a residual network. However, these models function as ``black boxes'' and impose a significant computational burden, typically requiring high-end GPUs and large datasets for training. 

\textbf{Contribution Positioning.} Despite their success, current literature lacks a balance between interpretability and efficiency. The proposed \ac{PRISM} framework bridges this gap by focusing on the perinuclear interface through a multiscale zonal approach. Unlike \ac{DL} counterparts, \ac{PRISM} is a low-cost, GPU-less methodology that provides transparency via verifiable biological markers. By avoiding exhaustive membrane delineation and dense neural computations, it provides an alternative with minimal computational overhead, making it ideal for deployment on resource-constrained diagnostic devices.

\section{Material and Methods}\label{sec:method}

\begin{figure}[ht]
    \centering
    \includegraphics[width=\linewidth]{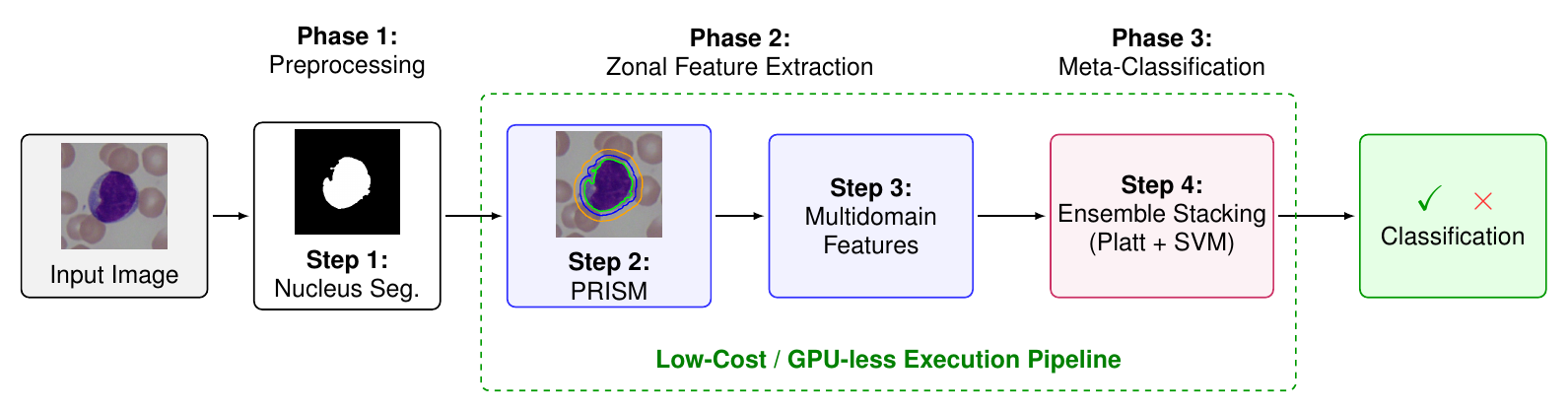}
    \caption{Steps of Proposed Method.}
    \label{fig:method}
\end{figure}

The proposed \ac{PRISM} is a lightweight approach designed for high-throughput \ac{ALL} screening without GPU dependency. The methodology, illustrated in Figure~\ref{fig:method}, is organized into three main phases encompassing four integrated steps: (i) preprocessing via nucleus segmentation, (ii) \ac{PRISM}-based spatial mapping, (iii) multidomain feature extraction, and (iv) meta-classification using calibrated ensemble stacking.

\subsection{Dataset}

\begin{figure}[ht]
    \centering
    \includegraphics[width=\textwidth]{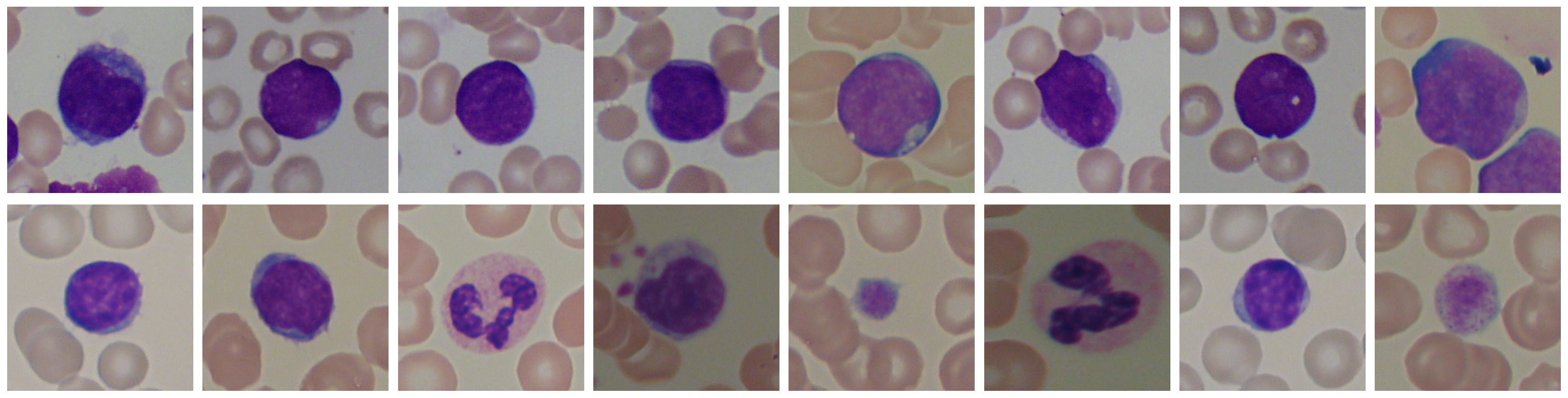}
    \caption{Samples from ALL-IDB2 dataset. Top Row: Pathogenic lymphoblasts. Bottom Row: Healthy lymphocytes.}
    \label{fig:dataset_samples}
\end{figure}

The proposed \ac{PRISM} method was evaluated using the \ac{ALL}-IDB2 dataset\footnote{\url{https://scotti.di.unimi.it/all/}} \cite{Labati2011}, a publicly available benchmark expert-annotated for automated hematological screening. Specifically tailored for single-cell classification tasks, this subset comprises $260$ isolated cell images, perfectly balanced between $130$ pathogenic leukemic lymphoblasts and $130$ healthy lymphocytes (Figure~\ref{fig:dataset_samples}).

\subsection{Pre-processing and Nuclear Segmentation}

To mitigate the illumination and staining inconsistencies inherent to blood smear imaging, the input images were standardized ($256 \times 256$ pixels) and mapped to the CIELAB color space. The structural definition was enhanced by applying contrast-limited adaptive histogram equalization (CLAHE) exclusively to the lightness ($L$) channel. 

Exploiting the basophilic nature of leukemic nuclei, images were projected into CIELAB and HSV spaces to compute a custom chromatic scoring map, $S_{chroma}$. This function maximizes the nucleocytoplasmic contrast by inversely weighting the blue-yellow opponent channel ($B$) and positively integrating saturation ($S$):
$$S_{chroma} = (255 - B) + 0.5 \times S$$

After Gaussian smoothing, Otsu's thresholding was 75\% crop to prevent background bias. For edge cases where extreme staining caused thresholding anomalies (yielding areas $<1\%$ or $>65\%$), a robust $K$-Means clustering fallback ($k=3$) in the CIELAB space autonomously isolated the darkest, adequately saturated cluster. The binary masks were refined using morphological opening, closing, and hole-filling operations.

Finally, topological heuristic filtering was used to discard artifacts and border-truncated cells. The primary nuclear mask ($M_n$) was identified by maximizing a multi-objective fitness function ($\mathcal{F}$) that balances solidity, circularity, and mean internal saturation ($\bar{S}$):
$$\mathcal{F} = 0.55 \times \text{Solidity} + 0.35 \times \text{Circularity} + 0.10 \times \left(\frac{\bar{S}}{255}\right)$$
This optimal component, $M_n$, serves as a geometric anchor for all subsequent spatial extractions.

\subsection{PRISM Approach}

\begin{figure*}[!ht]
    \centering
    \begin{tabular}{cccc}
        % Títulos das Colunas
        Original  & 
        Nucleus Mask ($M_n$) & 
        Zonal Isolation & 
        PRISM Overlay \\
        
        % Linha 1: Cancer 1
        \includegraphics[width=0.2\textwidth]{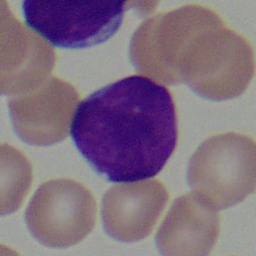} &
        \includegraphics[width=0.2\textwidth]{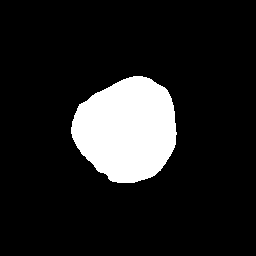} &
        \includegraphics[width=0.2\textwidth]{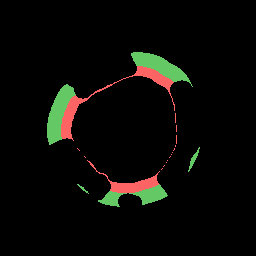} &
        \includegraphics[width=0.2\textwidth]{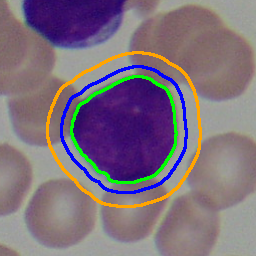} \\
        % Linha 2: Saudável 2
        \includegraphics[width=0.2\textwidth]{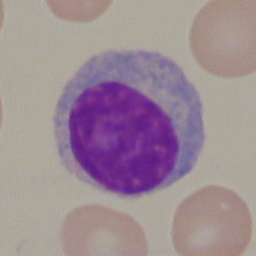} &
        \includegraphics[width=0.2\textwidth]{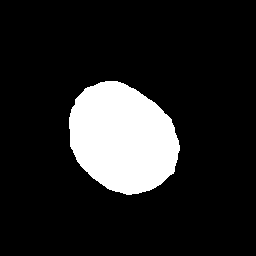} &
        \includegraphics[width=0.2\textwidth]{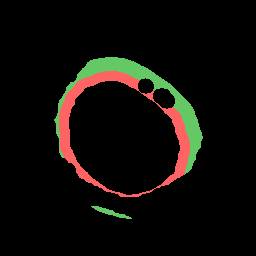} &
        \includegraphics[width=0.2\textwidth]{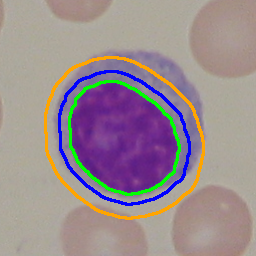} \\
    \end{tabular}
    \caption{Validation of the PRISM multiscale spatial decomposition across highly pleomorphic hematological profiles. Top row: Pathogenic lymphoblast \ac{ALL}. Bottom row: Healthy lymphocyte.}
    \label{fig:prism_pipeline}
\end{figure*}

The main contribution of this study is the \ac{PRISM} approach (Algorithm \ref{alg:prism}), which replaces error-prone cell membrane delineation with a multiscale concentric sampling strategy anchored on $M_n$ (Figure \ref{fig:prism_pipeline}). To handle morphological variance, \ac{PRISM} employs an adaptive dilation. Based on the nuclear area $A_n$ and its equivalent radius $R_{eq} = \sqrt{A_n/\pi}$, the expansion radius $\delta^*$ is dynamically bounded to prevent overextension: $\delta^* = \min(d, \max(6, \lfloor 1.6 \times R_{eq} \rfloor))$, where $d \in \{10, 24\}$ represents the empirical offsets.

Let $\mathcal{D}_{\delta^*}(M_n)$ denote the morphological dilation with an elliptical structuring element. To isolate the cytoplasmic signatures, two mutually exclusive spatial zones are generated via set difference operations and constrained by a gross cell boundary approximation ($\mathcal{C}$): the proximal interface $Z_1 = (\mathcal{D}_{\delta_1^*}(M_n) \setminus M_n) \cap \mathcal{C}$ and the distal cytoplasmic region $Z_2 = (\mathcal{D}_{\delta_2^*}(M_n) \setminus \mathcal{D}_{\delta_1^*}(M_n)) \cap \mathcal{C}$. These domains ($M_n, Z_1, Z_2$) serve as independent spatial bases for the localized extraction of morphological, chromatic, and textural biomarkers.

\begin{algorithm}[ht]
\footnotesize
\caption{PRISM: Multiscale Zonal Extraction Pipeline.}
\label{alg:prism}
\begin{algorithmic}[1]
\REQUIRE Enhanced Image $I$, Nuclear Mask $M_n$, Cell Boundary $\mathcal{C}$, Base Radii $d_1, d_2$
\ENSURE Multidimensional Feature Vector $V_f$

\STATE \COMMENT{\textbf{Step 1: Adaptive Spatial Decomposition}}
\STATE $R_{eq} \leftarrow \sqrt{\text{Area}(M_n) / \pi}$
\FOR{$i \in \{1, 2\}$}
    \STATE $\delta_i^* \leftarrow \min(d_i, \max(6, \lfloor 1.6 \times R_{eq} \rfloor))$ \COMMENT{Dynamic dilation bounds}
    \STATE $E_i \leftarrow \text{Dilate}(M_n, \text{radius}=\delta_i^*)$
\ENDFOR
\STATE $Z_1 \leftarrow (E_1 \setminus M_n) \cap \mathcal{C}$ \COMMENT{Proximal Zone}
\STATE $Z_2 \leftarrow (E_2 \setminus E_1) \cap \mathcal{C}$ \COMMENT{Distal Zone}

\STATE \COMMENT{\textbf{Step 2: Feature Engineering \& Gradient Computation}}
\STATE $V_f \leftarrow \emptyset$
\FOR{\textbf{each} spatial domain $\mathcal{S} \in \{M_n, Z_1, Z_2, \mathcal{C}\}$}
    \STATE $f_{\mathcal{S}} \leftarrow \text{Extract}(I, \mathcal{S})$ \COMMENT{Morphology, Color Stats, GLCM, LBP}
    \STATE $V_f \leftarrow V_f \cup f_{\mathcal{S}}$
\ENDFOR
\STATE $\Delta \leftarrow \text{ComputeGradients}(M_n, Z_1, Z_2)$ \COMMENT{Extract chromatic/textural decay}
\STATE $V_f \leftarrow V_f \cup \Delta$

\RETURN $V_f$
\end{algorithmic}
\end{algorithm}

\subsection{Feature Extraction and Characterization}

Quantitative descriptors were extracted from the defined topological domains ($M_n, Z_1, Z_2, \mathcal{C}$) to map leukemogenic alterations into a discriminative feature vector ($V_f$). 

\textbf{1. Morphological and Topological Descriptors:} To quantify nuclear hypertrophy and pleomorphism, geometric properties, including area, circularity, solidity, and boundary roughness, were computed. Topological ratios, specifically the nucleocytoplasmic (N:C) ratio defined as $Area(M_n) / Area(\mathcal{C})$, were calculated to characterize the proportional dominance of the nucleus, a critical clinical marker of \ac{ALL}.

\textbf{2. Chromatic Statistics and Spatial Gradients:} First-order statistical moments ($\mu, \sigma$) were extracted for the \ac{RGB} and grayscale channels. A key innovation of \ac{PRISM} is the computation of spatial intensity gradients ($\Delta$) across the nucleus-ring interfaces:
\begin{equation}
\Delta_{\mu}^{c} = \mu(M_n)_{c} - \mu(Z_i)_{c}; \quad \Delta_{\sigma}^{c} = \sigma(M_n)_{c} - \sigma(Z_i)_{c}
\end{equation}
These differential features encode chromatic decay from the nucleus to the cellular periphery, effectively capturing the staining affinity variations characteristic of lymphoblasts.

\textbf{3. Textural Analysis:} To model chromatin distribution, macro-textural properties (Contrast, Homogeneity, Energy, and Correlation) were computed via \ac{GLCM} for $M_n$ and zones $Z_i$, averaging results across distances $d \in \{1, 3\}$ and four angular orientations to ensure rotational invariance. In addition, micro-textural heterogeneities were captured using a standard 8-neighborhood \ac{LBP} operator. The resulting 256-bin histograms were downsampled to 32-bin representations via localized summation to mitigate dimensionality and prevent overfitting during the classification stage.

The concatenation of these multidomain attributes yields a robust and interpretable multidimensional vector ($V_f$) for each cell instance, bridging the gap between raw pixel data and clinical morphology.

\subsection{Ensemble Stacking and Probability Calibration}

To handle the heterogeneity of $V_f$, the framework employs a Heterogeneous Ensemble Stacking architecture. This meta-learning strategy combines the diverse inductive biases of multiple base learners to optimize the final decision boundary. The architecture is structured at two levels: Level-0 and Level-1. 

\textbf{Level-0} consists of a diverse pool of base classifiers: Tree-based ensembles \ac{RF}, \ac{ET}, and XGBoost, margin-based models (\ac{SVM}), and distance-based algorithms (\ac{KNN}). For the distance-sensitive models, the input space was standardized using Z-score normalization. To ensure that the meta-classifier receives reliable inputs, a Probability Calibration stage was integrated. Because algorithms such as the \ac{SVM} and \ac{RF} often produce uncalibrated scores, Platt Scaling (sigmoid calibration) was applied via 5-fold cross-validation. This procedure ensures that the Level-0 outputs represent the true posterior probabilities rather than raw decision scores, providing a statistically sound and normalized input space for the meta-learning stage.

\textbf{Level-1 Meta-Classifier}, implemented as a \ac{SVM} with a \ac{RBF} kernel, was trained using an \ac{OOF} strategy to prevent data leakage. During training, 5-fold Stratified Cross-Validation was used to generate OOF probabilities, forming a lower-dimensional meta-feature space. During inference, Level-0 models generate calibrated probabilities, which the Level-1 model synthesizes into the final diagnostic decision, maximizing generalization and mitigating overfitting.

\subsection{Evaluation Metrics}

All experiments were conducted using a 5-fold Stratified Cross-Validation scheme. For reproducibility, a fixed random seed was applied globally to all stochastic processes, including dataset partitioning and model initialization. In the context of automated oncological screening, standard accuracy is a limited metric. Consequently, the performance was quantified using clinical indicators: sensitivity (True Positive Rate) and specificity (True Negative Rate), which assess the model's reliability in identifying pathological and healthy samples, respectively.

In addition, we used the \ac{MCC}, defined in Equation \ref{eq:mcc}, a statistical rate for binary classification, as it produces a high score only if the model achieves balanced results across all four confusion matrix categories (True Positives, True Negatives, False Positives, and False Negatives).

\begin{equation}\label{eq:mcc}
MCC = \frac{TP \times TN - FP \times FN}{\sqrt{(TP+FP)(TP+FN)(TN+FP)(TN+FN)}}
\end{equation}

\section{Results and Discussion}\label{sec:results_and_discussion}  

All experiments were conducted in a virtual machine with 32GB of RAM, 32\textit{v}CPU, and an NVIDIA Quadro RTX~6000 GPU on the Fabric testbed~\cite{fabric-2019}. The experiments were programmed using Python~3.12 and PyTorch~3.10. The implementation and all experimental results are publicly available in the GitHub repository\footnote{Available at: \url{https://github.com/larissafrodrigues/prism-all}}.

\subsection{Performance of the PRISM}
To evaluate the discriminative capacity of the multiscale zonal features extracted by the \ac{PRISM}, an exhaustive search evaluating all $63$ possible combinatorial configurations of the base classifiers ($k \in [1, 6]$) was conducted. Base classifiers, comprising \ac{RF}, \ac{ET}, \ac{SVM}, \ac{LogReg}, \ac{KNN}, and XGBoost, were evaluated as standalone models and in complex multi-model architectures. All stacking combinations utilized a calibrated RBF SVM as the Level-1 meta-classifier.

The experimental results validate the robustness of the morphological, chromatic, and textural signatures extracted. Remarkably, ten distinct stacking configurations achieved the maximum macro-accuracy of $98.46\%$ ($\text{MCC} = 0.9698$), demonstrating that the spatial and gradient features inherently mapped the biological nuances of the leukemic domain, independent of a specific algorithm bias.

\begin{table}[htbp]
\centering
\caption{Performance Metrics of the Top-Performing Ensemble Stacking Architectures}
\label{tab:results}
 \renewcommand{\arraystretch}{1.1} 
\resizebox{\columnwidth}{!}{%
\begin{tabular}{lccccc}
\hline
\textbf{Level-0 Base Models} & \textbf{Accuracy} & \textbf{Balanced Acc.} & \textbf{MCC} & \textbf{AUC-ROC} & \textbf{PR-AUC} \\ \hline
ExtraTrees + SVM + LogReg & \textbf{0.9846} & \textbf{0.9846} & \textbf{0.9698} & \textbf{0.9896} & \textbf{0.9937} \\
RF + SVM + LogReg         & \textbf{0.9846} & \textbf{0.9846} & \textbf{0.9698} & 0.9822 & 0.9720 \\
RF + LogReg               & \textbf{0.9846} & \textbf{0.9846} & \textbf{0.9698} & 0.9816 & 0.9714 \\
ExtraTrees + RF + SVM     & \textbf{0.9846} & \textbf{0.9846} & \textbf{0.9698} & 0.9813 & 0.9714 \\
RF + SVM                  & \textbf{0.9846} & \textbf{0.9846} & \textbf{0.9698} & 0.9807 & 0.9712 \\ \hline
\end{tabular}%
}
\end{table}

As detailed in Table \ref{tab:results}, while multiple configurations converged to the optimal diagnostic accuracy, the ensemble comprising Extra Trees, SVM, and Logistic Regression (\textit{ET+SVM+LogReg}) emerged as the superior architecture. This specific trio yielded the highest probabilistic stability, achieving a threshold-independent AUC-ROC of $0.9896$ and a Precision-Recall AUC (PR-AUC) of $0.9937$. The near-perfect PR-AUC is of paramount clinical importance because it indicates an exceptionally low false discovery rate within the minority pathogenic class (lymphoblasts).

\subsection{Ablation Study: The Impact of Ensemble Dimensionality}
To contextualize the performance gains achieved by the stacking architecture, an ablation study was performed across different ensemble dimensionalities ($k$). For conciseness, Table \ref{tab:ablation_results} summarizes the optimal configurations for each value of $k$, comparing them against the single base classifier baselines (the complete combinatorial results for all 63 configurations are provided at the {\footnotesize\url{https://github.com/larissafrodrigues/prism-all}}).

\begin{figure*}[!ht]
    \centering
    \begin{minipage}{0.48\textwidth}
        \centering
        \includegraphics[width=\linewidth]{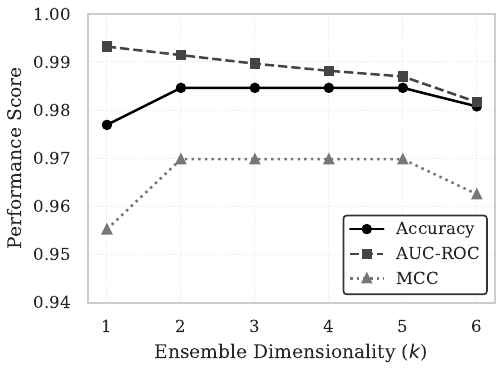}
        \caption{Ablation analysis.}
        \label{fig:ablation_plateau}
    \end{minipage}\hfill
    \begin{minipage}{0.48\textwidth}
        \centering
        \includegraphics[width=\linewidth]{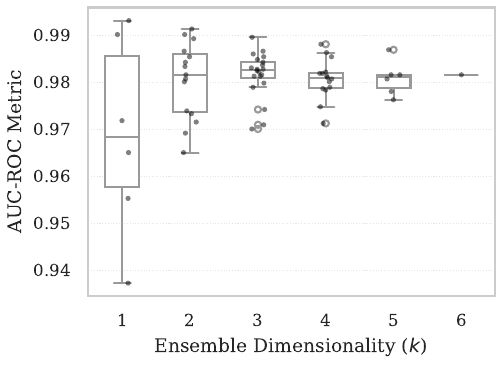}
        \caption{AUC-ROC scores.}
        \label{fig:auc_distribution}
    \end{minipage}
\end{figure*}

The data revealed a clear performance plateau, as illustrated in Figure \ref{fig:ablation_plateau}. While single classifiers (e.g., Extra Trees) achieved a baseline accuracy of $97.69\%$, the integration of the meta-classifier was critical for crossing the $98\%$ threshold. Fusing merely two models (\textit{RF + SVM}) immediately maximized the hard-label metrics ($98.46\%$ accuracy). Expanding the ensemble to three models optimized the soft-probability metrics (AUC). 

However, further increasing the ensemble complexity ($k \ge 4$) yielded no additional accuracy gains, and the complete heavyweight ensemble ($k=6$) exhibited a slight degradation in performance ($98.07\%$ accuracy). This slight performance drop in overly complex ensembles suggests the onset of dimensional overfitting at the meta-classifier level, a phenomenon further corroborated by the variance in the AUC-ROC distribution shown in Figure \ref{fig:auc_distribution}.

\begin{table}[!ht]
\centering
\caption{Comparative Performance of Single Classifiers vs Optimal Stacking Ensembles}
\label{tab:ablation_results}
 \renewcommand{\arraystretch}{1.1} 
\resizebox{\textwidth}{!}{%
\begin{tabular}{clccccc}
\hline
\textbf{\begin{tabular}[c]{@{}c@{}}Ensemble\\ Size (k)\end{tabular}} & \textbf{Model Configuration} & \textbf{Accuracy} & \textbf{Balanced Acc.} & \textbf{MCC} & \textbf{AUC-ROC} & \textbf{PR-AUC} \\ \hline
\multicolumn{7}{c}{\textit{Single Base Classifiers (Level-0 Baselines)}} \\ \hline
$k=1$ & Extra Trees (ET) & 0.9769 & 0.9769 & 0.9552 & 0.9902 & 0.9937 \\
$k=1$ & Random Forest (RF) & 0.9730 & 0.9730 & 0.9478 & 0.9852 & 0.9876 \\
$k=1$ & Support Vector Machine (SVM) & 0.9653 & 0.9653 & 0.9334 & 0.9893 & 0.9936 \\ \hline
\multicolumn{7}{c}{\textit{Optimal Ensemble Stacking Configurations (Level-1 Meta: SVM)}} \\ \hline
$k=2$ & RF + SVM & \textbf{0.9846} & \textbf{0.9846} & \textbf{0.9698} & 0.9807 & 0.9712 \\
$k=3$ & ET + SVM + LogReg & \textbf{0.9846} & \textbf{0.9846} & \textbf{0.9698} & \textbf{0.9896} & \textbf{0.9937} \\
$k=4$ & ET + RF + SVM + LogReg & \textbf{0.9846} & \textbf{0.9846} & \textbf{0.9698} & 0.9813 & 0.9714 \\
$k=5$ & ET + RF + SVM + KNN + LogReg & \textbf{0.9846} & \textbf{0.9846} & \textbf{0.9698} & 0.9816 & 0.9751 \\ \hline
\multicolumn{7}{c}{\textit{Complete Heavyweight Ensemble}} \\ \hline
$k=6$ & All Base Classifiers Combined & 0.9807 & 0.9807 & 0.9625 & 0.9816 & 0.9754 \\ \hline
\end{tabular}%
}
\end{table}

\subsection{Interpretability and Computational Efficiency}

Beyond predictive metrics, the \ac{PRISM} offers two key diagnostic advantages: using distinct spatial domains ($M_n$, $Z_1$, $Z_2$) ensures classification based on interpretable spatial gradients, contrasting with the opaque ``black-box'' nature of end-to-end \ac{DL} paradigms.

In addition, the ablation study proves that a low-latency configuration utilizing merely two base classifiers (\textit{RF + SVM}) is sufficient to preserve the $98.46\%$ diagnostic accuracy while drastically minimizing the computational overhead. This demonstrates that the \ac{PRISM} framework effectively translates the computationally expensive task of full-cell segmentation into a lightweight, high-performance tabular learning problem, confirming its suitability for rapid clinical screening in resource-constrained environments.

\subsection{Comparison with DL Baselines}

In contemporary medical image analysis, end-to-end \ac{DL} models, particularly \acp{CNN} and \acp{ViT}, are frequently used as the default computational paradigms. To evaluate the \ac{PRISM} approach, a comparative analysis was conducted with five widely adopted state-of-the-art \ac{DL} architectures. 

The baseline models included classical architectures (ResNet-18 and VGG16), a dense-connectivity model (DenseNet-121), a highly optimized modern \ac{CNN} (EfficientNet-B0), and a state-of-the-art attention-based model (Swin Transformer - Swin-T). All \ac{DL} models were trained with Adam optimizer (learning rate =$1\times10^{-4}$, weight decay=$1\times10^{-4}$) and binary cross-entropy with logits for 30 epochs, using a batch size of 16. All models were initialized with ImageNet pre-trained weights and fine-tuned on the raw, unsegmented $256 \times 256$ ALL-IDB2 images using the same stratified 5-fold cross-validation scheme to ensure a fair comparison of the results.

As shown in Table \ref{tab:cnn_comparison}, the \ac{PRISM}-driven ensemble outperformed all the \ac{DL} baselines in terms of accuracy and \ac{MCC}. 

\begin{table}[!ht]
\centering
\caption{Comparison between PRISM and \ac{DL} Architectures}
\label{tab:cnn_comparison}
\resizebox{\columnwidth}{!}{%
\begin{tabular}{llcc}
\hline
\textbf{Methodology} & \textbf{Computational Paradigm} & \textbf{Accuracy} & \textbf{MCC} \\ \hline
Swin Transformer (Swin-T) & Vision Transformer (Self-Attention) & 0.9192 & 0.8597 \\
ResNet-18                 & CNN        & 0.9654 & 0.9333 \\
DenseNet-121              & CNN        & 0.9769 & 0.9557 \\
VGG16                     & CNN      & 0.9769 & 0.9550 \\
EfficientNet-B0           & CNN       & 0.9808 & 0.9630 \\ \hline
\textbf{PRISM (ET+SVM+LogReg)} & \textbf{Multiscale Zonal Extraction + Stacking} & \textbf{0.9846} & \textbf{0.9698} \\ \hline
\end{tabular}%
}
\end{table}

EfficientNet-B0 achieved the highest performance ($98.08\%$ accuracy) among \ac{DL} models owing to its optimal network balance. Swin-T showed lower performance ($91.92\%$ accuracy), typical of self-attention mechanisms that require massive datasets to generalize properly in hematological imaging.

The superiority of the \ac{PRISM} over these \ac{DL} paradigms underscores its clinical and computational advantages.
\begin{enumerate}
    \item \textbf{Biological Focus vs. Background Noise:} End-to-end CNNs processing full-frame smears are highly susceptible to learning confounding background artifacts (e.g., erythrocyte density, staining dye variations). \ac{PRISM} geometrically forces the classifier to evaluate only the relevant biological transition between the nucleus and the proximal/distal cytoplasm.
    \item \textbf{Interpretability and Hardware Efficiency:} While modern architectures like DenseNet-121 and EfficientNet-B0 act as opaque ``black boxes'' demanding energy-intensive GPU clusters for training and inference, \ac{PRISM} effectively reduces the high-dimensional image processing task into a lightweight, tabular Machine Learning problem.
\end{enumerate}

By combining targeted spatial morphology with calibrated ensemble stacking, the \ac{PRISM} methodology establishes a highly accurate, explainable, and resource-efficient standard for automated lymphoblast screening, outperforming current \ac{DL} paradigms.

\subsection{Comparison with State-of-the-Art}

To benchmark the proposed \ac{PRISM} approach, its classification performance was evaluated against previous studies that used the ALL-IDB2 dataset. The comparative analysis focusing on accuracy (ACC) is summarized in Table~\ref{tab:comparison}.

\begin{table}[ht]
\centering
\caption{Comparison with previous studies.}
\label{tab:comparison}
\resizebox{\columnwidth}{!}{%
\begin{tabular}{@{}llc@{}}
\toprule
\textbf{Author} & \textbf{Methodology} & \textbf{ACC (\%)} \\ \midrule
\cite{Rodrigues2016} & Hybrid CNN (ResNet + GA) & 85.00 \\
\cite{Sahlol2017} & Handcrafted (EOF+MLP) & 91.80 \\
\cite{Faria2018} & Handcrafted (SIFT+SURF) & 97.22 \\
\cite{Sahlol2019} & Handcrafted (KNN) & 95.67 \\
\cite{Sahlol2020} & CNN (VGG + SESSA) & 96.11 \\
\cite{Das2021} & CNN (MobileNet + ResNet) & 97.18 \\
\cite{Rodrigues2022} & CNN (ResNet + GA) & 98.46 \\
\cite{Das2025} & Hybrid CNN Ensemble (MobileNet + ShuffleNet) & 98.46 \\
\cite{Hita2026} & Bayesian CNN + Augmentation & 98.65 \\ \midrule
\textbf{PRISM (Ours)} & \textbf{Zonal Stacking Ensemble} & \textbf{98.46} \\ \bottomrule
\end{tabular}%
}
\end{table}

The results demonstrate that \ac{PRISM} achieves a highly accuracy of 98.46\%, matching the performance of complex \ac{DL} ensembles \cite{Rodrigues2022, Das2025} and significantly outperforming traditional handcrafted methods \cite{Faria2018, Sahlol2019}. Although the Bayesian \ac{CNN} approach by \cite{Hita2026} reports a higher accuracy, it relies heavily on data augmentation and computationally expensive GPU-accelerated training.

In contrast, \ac{PRISM} provides a low-cost, GPU-less alternative. By leveraging localized spatial domains ($Z_1, Z_2$) coupled with an \ac{OOF} stacking meta-classifier (\ac{SVM}), our framework extracts highly discriminative signatures directly from the cell morphology. This eliminates the need for deep feature extraction, extensive transfer learning, and exhaustive hyperparameter tuning. Consequently, \ac{PRISM} reduces training times and operates efficiently in standard CPU environments, establishing itself as a highly scalable solution for resource-constrained clinical screenings.

\section{Conclusion}\label{sec:conclusion}

Accurate automated screening of \ac{ALL} is often hindered by low contrast, cytoplasmic variability, and the brittleness of full-cell membrane segmentation. This study presents \ac{PRISM}, an interpretable framework that leverages stable nuclear geometry and extracts chromatic and textural cues across adaptive perinuclear zones to capture the morphology linked to leukemic transformation. Combined with a calibrated stacking ensemble, \ac{PRISM} achieved a strong performance on ALL-IDB2, reaching an accuracy of 98.46\% and an \ac{MCC} of 0.9698.

Comparative analyses indicate that \ac{PRISM} performs on par with and, in several cases, surpasses contemporary \ac{DL} models while remaining lightweight and transparent, enabling fast inference without specialized hardware in resource-limited or decentralized settings. Future studies will integrate \ac{PRISM} into portable or edge-based microscopy pipelines to support accessible point-of-care diagnostics. In addition, we plan to validate additional datasets, such as real-world clinical data from different laboratories. Overall, the results show that segmentation-light, interpretable representations can deliver high-accuracy ALL screening with practical efficiency and clearer decision-making support.

\section*{Acknowledgments}

The authors gratefully acknowledges the financial support of FAPEMIG (Grant \#APQ00923-24). Andr\'e R. Backes gratefully acknowledges the financial support of CNPq (Grant \#302790/2024-1). 

\bibliographystyle{sbc}
\bibliography{references}

@INPROCEEDINGS{Labati2011,
  author={Labati, Ruggero Donida and Piuri, Vincenzo and Scotti, Fabio},
  booktitle={2011 18th IEEE International Conference on Image Processing}, 
  title={All-IDB: The acute lymphoblastic leukemia image database for image processing}, 
  year={2011},
  volume={},
  number={},
  pages={2045-2048},
  doi={10.1109/ICIP.2011.6115881}}

@INPROCEEDINGS{Scotti2005,
  author={Scotti, F.},
  booktitle={CIMSA. 2005 IEEE International Conference on Computational Intelligence for Measurement Systems and Applications, 2005.}, 
  title={Automatic morphological analysis for acute leukemia identification in peripheral blood microscope images}, 
  year={2005},
  volume={},
  number={},
  pages={96-101},
  doi={10.1109/CIMSA.2005.1522835}}

@inproceedings{Rodrigues2016,
  title={Leukocytes classification in microscopy images for acute lymphoblastic leukemia identification},
  author={Rodrigues, Larissa Ferreira and Silva, Jonathan Henrique and Gondim, Pedro Henrique Cunha Campos and Mari, Joao Fernando},
  booktitle={XII Workshop de Vis{\~a}o Computacional},
  pages={68--73},
  year={2016}
}

@inproceedings{Faria2018,
  title={Cell classification using handcrafted features and bag of visual words},
  author={de Faria, Lucas Costa and Rodrigues, Larissa Ferreira and Mari, Joao Fernando},
  booktitle={Anais do XIV Workshop de Vis{\~a}o Computacional},
  pages={68--73},
  year={2018}
}

@Article{Sahlol2019,
author={Sahlol, Ahmed T.
and Abdeldaim, Ahmed M.
and Hassanien, Aboul Ella},
title={Automatic acute lymphoblastic leukemia classification model using social spider optimization algorithm},
journal={Soft Computing},
year={2019},
month={Aug},
day={01},
volume={23},
number={15},
pages={6345-6360},
issn={1433-7479},
doi={10.1007/s00500-018-3288-5},
url={https://doi.org/10.1007/s00500-018-3288-5}
}

@article{fabric-2019, 
title={{FABRIC: A national-scale programmable experimental network infrastructure}},
author={Baldin, Ilya and Nikolich, Anita and Griffioen, James and Monga, Indermohan Inder S and Wang, Kuang-Ching and Lehman, Tom and Ruth, Paul}, 
journal={IEEE Internet Computing}, 
year={2019}, 
volume={23}, 
number={6}, 
pages={38--47}, 
publisher={IEEE} 
}

@INPROCEEDINGS{Sahlol2017,
  author={A. T. {Sahlol} and F. H. {Ismail} and A. {Abdeldaim} and A. E. {Hassanien}},
  booktitle={2017 12th International Conference on Computer Engineering and Systems (ICCES)}, 
  title={Elephant herd optimization with neural networks: A case study on acute Lymphoblastic Leukemia diagnosis}, 
  year={2017},
  volume={},
  number={},
  pages={657-662},
  doi={10.1109/ICCES.2017.8275387}}

@article{Das2021,
title = {An efficient deep Convolutional Neural Network based detection and classification of Acute Lymphoblastic Leukemia},
journal = {Expert Systems with Applications},
volume = {183},
pages = {115311},
year = {2021},
issn = {0957-4174},
doi = {https://doi.org/10.1016/j.eswa.2021.115311},
url = {https://www.sciencedirect.com/science/article/pii/S0957417421007405},
author = {Pradeep Kumar Das and Sukadev Meher}
}

@article{Das2025,
title = {An efficient deep learning system for automatic detection of Acute Lymphoblastic Leukemia},
journal = {ISA Transactions},
volume = {158},
pages = {488-496},
year = {2025},
issn = {0019-0578},
doi = {https://doi.org/10.1016/j.isatra.2024.12.043},
url = {https://www.sciencedirect.com/science/article/pii/S001905782400627X},
author = {Pradeep Kumar Das and Sukadev Meher and Adyasha Rath and Ganapati Panda}
}

@Article{Sahlol2020,
author={Sahlol, Ahmed T.
and Kollmannsberger, Philip
and Ewees, Ahmed A.},
title={Efficient Classification of White Blood Cell Leukemia with Improved Swarm Optimization of Deep Features},
journal={Scientific Reports},
year={2020},
month={Feb},
day={13},
volume={10},
number={1},
pages={2536},
issn={2045-2322},
doi={10.1038/s41598-020-59215-9},
url={https://doi.org/10.1038/s41598-020-59215-9}
}

@Article{Rodrigues2022,
author={Rodrigues, Larissa Ferreira
and Backes, Andr{\'e} Ricardo
and Traven{\c{c}}olo, Bruno Augusto Nassif
and de Oliveira, Gina Maira Barbosa},
title={Optimizing a Deep Residual Neural Network with Genetic Algorithm for Acute Lymphoblastic Leukemia Classification},
journal={Journal of Digital Imaging},
year={2022},
month={Jun},
day={01},
volume={35},
number={3},
pages={623-637},
issn={1618-727X},
doi={10.1007/s10278-022-00600-3},
url={https://doi.org/10.1007/s10278-022-00600-3}
}

@inproceedings{Dias2021,
 author = {Domingos {Dias Júnior} and Luana Cruz and João Diniz and Geraldo Braz Júnior and Aristófanes Silva},
 title = {Classificação automática de glóbulos brancos usando descritores de forma e textura e eXtreme Gradient Boosting},
 booktitle = {Anais do XXI Simpósio Brasileiro de Computação Aplicada à Saúde},
 location = {Evento Online},
 year = {2021},
 issn = {2763-8952},
 pages = {95--106},
 publisher = {SBC},
 address = {Porto Alegre, RS, Brasil},
 doi = {10.5753/sbcas.2021.16056},
 url = {https://sol.sbc.org.br/index.php/sbcas/article/view/16056}
}

@article{Hita2026,
title = {Reliable leukemia detection via transfer-enhanced Bayesian CNNs},
journal = {Computers in Biology and Medicine},
volume = {202},
pages = {111419},
year = {2026},
issn = {0010-4825},
doi = {https://doi.org/10.1016/j.compbiomed.2025.111419},
url = {https://www.sciencedirect.com/science/article/pii/S0010482525017731},
author = {Xhesina Hita and Farrukh Javed and Stefano Lodi}
}

@Article{Kizi2025,
AUTHOR = {Oybek Kizi, Rakhmonalieva Farangis and Theodore Armand, Tagne Poupi and Kim, Hee-Cheol},
TITLE = {A Review of Deep Learning Techniques for Leukemia Cancer Classification Based on Blood Smear Images},
JOURNAL = {Applied Biosciences},
VOLUME = {4},
YEAR = {2025},
NUMBER = {1},
ARTICLE-NUMBER = {9},
URL = {https://www.mdpi.com/2813-0464/4/1/9},
ISSN = {2813-0464},
DOI = {10.3390/applbiosci4010009}
}

@ARTICLE{Zolfaghari2022,
  title    = "A survey on automated detection and classification of acute
              leukemia and {WBCs} in microscopic blood cells",
  author   = "Zolfaghari, Mohammad and Sajedi, Hedieh",
  journal  = "Multimedia Tools and Applications",
  volume   =  81,
  number   =  5,
  pages    = "6723--6753",
  month    =  feb,
  year     =  2022
}

@article{Aby2024,
title = {A review on leukemia detection and classification using Artificial Intelligence-based techniques},
journal = {Computers and Electrical Engineering},
volume = {118},
pages = {109446},
year = {2024},
issn = {0045-7906},
doi = {https://doi.org/10.1016/j.compeleceng.2024.109446},
url = {https://www.sciencedirect.com/science/article/pii/S0045790624003732},
author = {Aswathy Elma Aby and S. Salaji and K.K. Anilkumar and Tintu Rajan}
}

@article{RodriguesMoreira2025,
title = {Deep learning based image classification for embedded devices: A systematic review},
journal = {Neurocomputing},
volume = {623},
pages = {129402},
year = {2025},
issn = {0925-2312},
doi = {https://doi.org/10.1016/j.neucom.2025.129402},
url = {https://www.sciencedirect.com/science/article/pii/S0925231225000748},
author = {Larissa Ferreira {Rodrigues Moreira} and Rodrigo Moreira and Bruno Augusto Nassif Travençolo and André Ricardo Backes}
}

\end{document}